\documentclass[runningheads]{llncs}
\usepackage[T1]{fontenc}
\usepackage{graphicx}
\usepackage{hyperref}
\usepackage{color}

\usepackage{amssymb}
\usepackage{silence}
\usepackage{amsmath} 
\usepackage{xcolor}
\usepackage{multirow}
\usepackage{tikz}
\usepackage{pgfplots}
\usepackage{tabularx} 
\usepackage{booktabs}
\usepackage[table]{xcolor} 
\usetikzlibrary{arrows.meta,positioning}
\usepgfplotslibrary{groupplots}
\pgfplotsset{compat=1.18}

\begin{document}
\title{QAPruner: Quantization-Aware Vision Token Pruning for Multimodal Large Language Models}
\titlerunning{QAPruner for Multimodal Large Language Models}
%
\author{Xinhao Wang \and
Zhonyu Xia \and
Zhiwei Lin \and
Zhe Li \and
Yongtao Wang\thanks{Corresponding author.}}
%
%
\institute{Wangxuan Institute of Computer Technology, Peking University \\
\email{\{wangxinhao, xiazhongyu, zwlin, wyt\}@pku.edu.cn} }

\maketitle              
\begin{abstract}

Multimodal Large Language Models (MLLMs) have shown strong reasoning ability, but their high computational and memory costs hinder deployment in resource-constrained settings. While Post-Training Quantization (PTQ) and vision token pruning are standard compression techniques, they are usually treated as independent optimizations. In this paper, we show that these two techniques are strongly coupled: naively applying semantic-based token pruning to PTQ-optimized MLLMs can discard activation outliers that are important for numerical stability and thus worsen quantization errors in low-bit regimes (\textit{e.g.}, W4A4). To address this issue, we propose a quantization-aware vision token pruning framework. Our method introduces a lightweight hybrid sensitivity metric that combines simulated group-wise quantization error with outlier intensity. By combining this metric with standard semantic relevance scores, the method retains tokens that are both semantically informative and robust to quantization. Experiments on standard LLaVA architectures show that our method consistently outperforms naive integration baselines. At an aggressive pruning ratio that retains only 12.5\% of visual tokens, our framework improves accuracy by 2.24\% over the baseline and even surpasses dense quantization without pruning. To the best of our knowledge, this is the first method that explicitly co-optimizes vision token pruning and PTQ for accurate low-bit MLLM inference.

\keywords{Multimodal large language models  \and Post-training quantization \and Vision token pruning.}
\end{abstract}
\section{Introduction}

Multimodal large language models (MLLMs) have recently emerged as a powerful paradigm for unified vision-language understanding and reasoning. By coupling a strong vision encoder with a large language model, systems such as LLaVA~\cite{llava,llavanext,liu2023improvedllava} enable open-ended visual question answering, instruction following, and multi-step reasoning over images. Despite their strong capabilities, the practical deployment of MLLMs is often constrained by high computational and memory costs arising from the large language backbone and dense patch tokens used for attention computation.
As MLLMs move from research prototypes to real-world services and edge devices, improving their efficiency while preserving accuracy becomes increasingly important.

Given the substantial training costs of MLLMs, optimization strategies that require minimal or no fine-tuning are highly advantageous. 
Two primary families of techniques have emerged as particularly effective. 
First, Post-Training Quantization (PTQ) compresses model weights and activations into low-bit formats (\textit{e.g.}, W4A4), significantly reducing memory bandwidth requirements and accelerating inference on compatible hardware. However, PTQ often compromises accuracy, particularly in low-bit regimes where activation outliers and distribution shifts induce significant quantization errors. Second, vision token pruning mitigates computational overhead by selectively retaining informative visual tokens, thereby alleviating sequence length constraints and reducing attention complexity. Nevertheless, pruning is inherently a lossy operation. Suboptimal selection criteria may lead to the removal of critical tokens, severely impairing vision-language performance.

\begin{figure}[t]
\centering
\resizebox{0.98\textwidth}{!}{\input{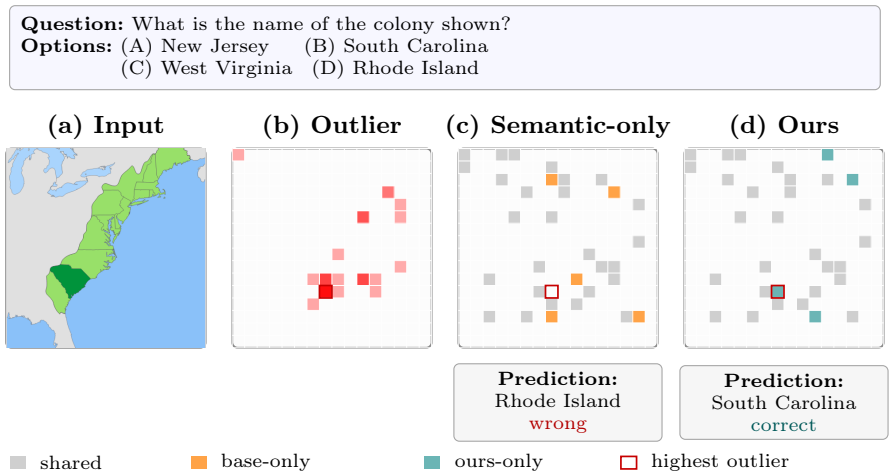}}
\vspace{-10pt}
\caption{Motivation teaser of quantization-aware vision token pruning on a real ScienceQA sample. Panel (a) shows the input image. Panel (b) shows the token-level outlier scores, with darker red cells indicating greater quantization sensitivity. Panel (c) shows the tokens kept by semantic-only pruning, which misses the highest-scoring outlier token and leads the quantized model to predict \emph{Rhode Island}. Panel (d) shows our selection, which preserves that token and recovers the correct answer \emph{South Carolina}.}
\vspace{-15pt}
\label{fig:motivation_teaser}
\end{figure}

An important observation is that quantization and token pruning, although often studied separately, interact strongly in MLLMs. Token pruning changes the activation distribution seen by the quantized model. As a result, quantization errors can increase substantially if the retained tokens fail to preserve numerically significant signals, especially those associated with outliers. Therefore, a token subset that is ``semantically relevant'' is not always ``quantization-robust''. Effective quantization may require retaining tokens that are important for numerical stability even when their direct semantic contribution appears limited. Figure~\ref{fig:motivation_teaser} illustrates this mismatch on a representative ScienceQA example: semantic-only pruning misses a highly quantization-sensitive token and yields an incorrect prediction, whereas our method preserves that token and recovers the correct answer. 
Existing pruning techniques are predominantly quantization-agnostic, often rendering them suboptimal for low-bit regimes (\textit{e.g.}, W4A4).
This observation motivates a unified framework that jointly considers pruning and quantization.

In this work, we propose a quantization-aware vision token pruning method tailored for PTQ-optimized MLLMs. 
%
%
To construct a more complete measure, we propose a hybrid sensitivity score that computes both the simulated group-wise error and the dynamic range for each visual token and combines them into a single score.
The dynamic range metric effectively identifies tokens with extreme outliers and captures global intensity, and the group-wise quantization error accurately simulates local detail loss during PTQ.
By integrating this quantization sensitivity metric with standard vision token pruning scores, our method prioritizes tokens that are difficult to quantize and contain important outliers. This design balances detail preservation and outlier protection, helping preserve critical visual information in the low-bit representation space.

Our primary insights and contributions are:

\begin{itemize}

\item \textbf{Pruning-Quantization Interdependence.} We show that vision token pruning and PTQ interact strongly in MLLMs. Semantic-only pruning can remove activation outliers that are important for numerical stability, which further worsens quantization errors in low-bit settings.

\item \textbf{Hybrid Quantization Sensitivity Metric.} We introduce a lightweight quantization sensitivity score to identify quantization-critical tokens. It combines dynamic range and simulated group-wise quantization error to capture both global outliers and local detail loss.

\item \textbf{Quantization-Aware Vision Token Pruning.} We propose a training-free pruning framework that integrates the hybrid sensitivity metric with semantic relevance scores. This design retains tokens that are both semantically informative and robust to quantization, enabling efficient low-bit inference.

\end{itemize}

\section{Related Works}

\subsection{Multimodal Large Language Models}

Recent advancements in Large Language Models (LLMs)~\cite{alayrac2022flamingo,grattafiori2024llama,touvron2023llama,touvron2023llama2} have catalyzed the development of MLLMs~\cite{llava,llavanext,liu2023improvedllava,qwenvl,InternVL,gemini2_5,gpt5,deepseekr1}, effectively extending the sophisticated reasoning capabilities of LLMs into the visual domain. 
Pioneering frameworks such as Flamingo~\cite{alayrac2022flamingo} showed that coupling frozen visual encoders with LLMs yields strong few-shot performance on vision-language tasks. 
The subsequent emergence of powerful open-source foundation models, notably the LLaMA series~\cite{grattafiori2024llama,touvron2023llama,touvron2023llama2}, further accelerated research in multimodal understanding. 
Using these pre-trained language backbones, researchers have developed instruction-tuned MLLMs~\cite{llava,liu2023improvedllava,llavanext} with strong visual reasoning ability. 
Architecturally, these models typically comprise a vision encoder (\textit{e.g.}, CLIP~\cite{clip}, SigLIP~\cite{siglip}) for visual feature extraction, a projection module to align visual and textual modalities, and a pre-trained LLM as the central reasoning engine. 
%
While early approaches primarily focused on basic cross-modal alignment, contemporary MLLMs have evolved to process high-resolution images, comprehend interleaved image-text inputs, and execute complex multimodal instructions. To further enhance fine-grained visual understanding, models such as Qwen-VL~\cite{qwenvl} and InternVL~\cite{InternVL} have scaled both their training corpora and input resolution capacities. 
Recent state-of-the-art models introduce even more sophisticated paradigms. 
%
Despite these remarkable advancements, the immense computational cost of deploying large-scale models remains a critical bottleneck, underscoring the urgent need for highly efficient inference techniques.

\subsection{Post Training Quantization}

To mitigate the substantial memory and computational demands of large-scale models, quantization has emerged as a pivotal compression technique~\cite{llmint8,SmoothQuant,awqquant,QuIP,GPTQ,OmniQuant}. 
Among various strategies, PTQ is particularly advantageous, as it compresses models to lower-precision formats (\textit{e.g.}, INT8 or INT4) without requiring extensive retraining or access to the full training corpus. 
Although PTQ has been successfully employed in LLMs to reduce weight and improve activation precision, extending it to MLLMs introduces unique challenges stemming from the disparate distributions of visual and textual features.

Recent studies have explored the application of PTQ across various components of MLLMs. 
For instance, MBQ~\cite{mbq} incorporates modality-specific sensitivity during the calibration phase to optimize quantization parameters. This approach preserves accuracy in low-bit regimes (\textit{e.g.}, W3 and W4A8) while achieving inference speedups through custom GPU kernels. 
Another framework, Q-VLM~\cite{qvlm}, optimizes cross-layer dependencies rather than relying on conventional sequential, layer-wise rounding. It utilizes activation entropy as a proxy to partition blocks and optimizes the visual encoder to disentangle these dependencies effectively.  
Furthermore, MQuant~\cite{mquant} addresses the high latency and distributional disparities between visual and textual tokens by introducing Modality-Specific Static Quantization (MSQ) and Attention-Invariant Flexible Switching (AIFS). These mechanisms assign distinct static scales to each modality and reorder tokens to circumvent computationally expensive online calculations. Concurrently, Rotation Magnitude Suppression (RMS) is employed to mitigate outliers arising from Hadamard transformations, ultimately delivering significant inference speedups with negligible accuracy degradation.

\subsection{Vision Tokens Pruning}

MLLMs' visual inputs are typically converted into token sequences that frequently become prohibitively long, particularly when processing high-resolution images~\cite{visionzip}. Long sequence length significantly increases the computational overhead of the LLM's attention mechanism, leading to substantial latency and memory consumption during inference. Vision token pruning mitigates this bottleneck by identifying and discarding redundant or uninformative visual tokens, thereby curtailing the effective sequence length processed by the language backbone.

Current pruning approaches range from fundamental, heuristic-based techniques to sophisticated strategies that leverage attention weights or gradient-based importance scores. 
%
For instance, LLaVA-PruMerge dynamically selects critical tokens based on class-token attention and subsequently merges similar tokens to preserve contextual information.
Similarly, TRIM~\cite{trim} optimizes LLaVA by computing text-vision token similarities and employing an outlier detection algorithm to rank and select salient tokens. 
VisionZip~\cite{visionzip} mitigates visual token redundancy by selectively retaining only the most informative tokens for subsequent language modeling. 
Furthermore, CDPruner~\cite{cdpruner} overcomes the limitations of the aforementioned methods by maximizing the conditional diversity of the retained visual tokens. By reformulating token selection as a Determinantal Point Process (DPP) conditioned on user instructions, CDPruner ensures that the selected subset provides a comprehensive visual representation that strictly aligns with the textual query.

At the same time, prior work on quantization has shown that activation outliers are important for preserving model accuracy~\cite{llmint8}. This suggests that token pruning methods guided solely by semantic relevance may discard tokens important for quantization robustness, a gap that existing pruning methods do not explicitly address.
Our framework addresses this gap by jointly optimizing quantization and vision token pruning to improve inference efficiency.

\section{Method}

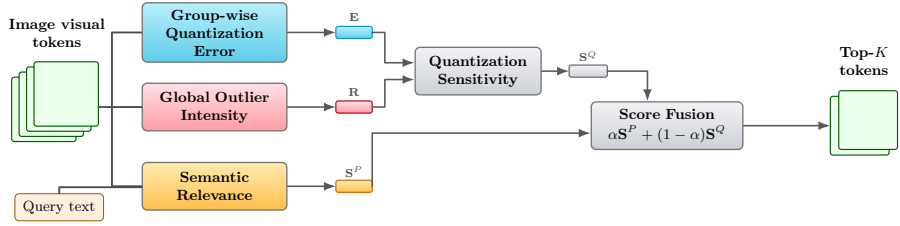
\begin{figure}[t]
\centering
\resizebox{0.98\textwidth}{!}{\begin{tikzpicture}[
    font=\footnotesize,
    >=Latex,
    line cap=round,
    line join=round,
    flow/.style={-Latex, draw=black!65, line width=0.95pt},
    plainflow/.style={draw=black!65, line width=0.95pt},
    trunk/.style={draw=black!70, line width=1.15pt},
    token/.style={rounded corners=1.6pt, draw=green!35!black, line width=0.7pt, fill=green!8},
    block/.style={
        rounded corners=3pt,
        draw=black!55,
        line width=0.8pt,
        minimum width=3.1cm,
        minimum height=1.02cm,
        align=center,
        inner sep=4pt
    },
    scorebar/.style={
        rounded corners=1.4pt,
        line width=0.7pt,
        minimum width=0.75cm,
        minimum height=0.24cm
    }
]

\definecolor{QAcyanA}{RGB}{184,239,251}
\definecolor{QAcyanB}{RGB}{95,205,232}
\definecolor{QApinkA}{RGB}{255,226,229}
\definecolor{QApinkB}{RGB}{255,155,165}
\definecolor{QAyellowA}{RGB}{255,235,176}
\definecolor{QAyellowB}{RGB}{255,190,77}
\definecolor{QAgrayA}{RGB}{242,243,246}
\definecolor{QAgrayB}{RGB}{205,208,214}
\definecolor{QAdark}{RGB}{67,76,89}

\foreach \dx/\dy in {0/0,0.16/0.11,0.32/0.22,0.48/0.33}{
    \draw[token] (\dx,\dy-0.69) rectangle ++(1.38,1.38);
}
\node[align=center, font=\bfseries] at (0.97,1.64) {Image visual\\tokens};

\node[
    rounded corners=2.5pt,
    draw=orange!65!black,
    fill=orange!12,
    minimum width=1.9cm,
    minimum height=0.58cm,
    inner sep=2pt
] (query) at (1.02,-2.10) {Query text};

\node[
    block,
    top color=QAcyanA,
    bottom color=QAcyanB,
    text width=2.8cm
] (qerr) at (4.35,1.65) {\textbf{Group-wise}\\\textbf{Quantization Error}};

\node[
    block,
    top color=QApinkA,
    bottom color=QApinkB,
    text width=2.8cm
] (outlier) at (4.35,0.05) {\textbf{Global Outlier}\\\textbf{Intensity}};

\node[
    block,
    top color=QAyellowA,
    bottom color=QAyellowB,
    text width=2.8cm
] (semantic) at (4.35,-1.65) {\textbf{Semantic}\\\textbf{Relevance}};

\node[
    block,
    top color=QAgrayA,
    bottom color=QAgrayB,
    text width=2.3cm,
    minimum width=2.7cm
] (qsense) at (10.00,0.82) {\textbf{Quantization}\\\textbf{Sensitivity}};

\node[
    block,
    top color=QAgrayA,
    bottom color=QAgrayB,
    text width=2.9cm,
    minimum width=3.25cm
] (fusion) at (14.05,-0.35) {\textbf{Score Fusion}\\[1pt]$\alpha \mathbf{S}^{P} + (1-\alpha)\mathbf{S}^{Q}$};

\foreach \dx/\dy in {0/0,0.18/0.12}{
    \draw[token] (17.55+\dx,-0.98+\dy) rectangle ++(1.26,1.26);
}
\node[align=center, font=\bfseries] at (18.27,1.04) {Top-$K$\\tokens};

\coordinate (split) at (2.15,0.05);
\draw[trunk] (1.86,0.05) -- (split);
\draw[trunk] (split) |- (qerr.west);
\draw[trunk] (split) -- (outlier.west);
\draw[trunk] (split) |- (semantic.west);
\draw[trunk] (query.north) |- ([yshift=-0.02cm]semantic.west);

\draw[scorebar, top color=QAcyanA, bottom color=QAcyanB, draw=cyan!70!black] (6.95,1.52) rectangle ++(0.78,0.26);
\node[font=\scriptsize\bfseries, text=QAdark] at (7.34,1.97) {$\mathbf{E}$};

\draw[scorebar, top color=QApinkA, bottom color=QApinkB, draw=red!70!black] (6.95,-0.08) rectangle ++(0.78,0.26);
\node[font=\scriptsize\bfseries, text=QAdark] at (7.34,0.37) {$\mathbf{R}$};

\draw[scorebar, top color=QAyellowA, bottom color=QAyellowB, draw=orange!80!black] (6.95,-1.78) rectangle ++(0.78,0.26);
\node[font=\scriptsize\bfseries, text=QAdark] at (7.34,-1.33) {$\mathbf{S}^{P}$};

\draw[scorebar, top color=QAgrayA, bottom color=QAgrayB, draw=gray!75!black] (11.95,0.69) rectangle ++(0.82,0.26);
\node[font=\scriptsize\bfseries, text=QAdark] at (12.36,1.14) {$\mathbf{S}^{Q}$};

\draw[flow] (qerr.east) -- (6.95,1.65);
\draw[flow] (outlier.east) -- (6.95,0.05);
\draw[flow] (semantic.east) -- (6.95,-1.65);

\coordinate (qsenseTopIn) at ([yshift=0.18cm]qsense.west);
\coordinate (qsenseBotIn) at ([yshift=-0.18cm]qsense.west);
\draw[flow] (7.73,1.65) -- (8.00,1.65) |- (qsenseTopIn);
\draw[flow] (7.73,0.05) -- (8.00,0.05) |- (qsenseBotIn);
\draw[flow] (qsense.east) -- (11.95,0.82);

\coordinate (fusionTopIn) at ([xshift=-0.42cm]fusion.north);
\coordinate (fusionWestIn) at ([yshift=-0.16cm]fusion.west);
\draw[flow] (12.77,0.82) -| (fusionTopIn);
\draw[flow] (7.73,-1.65) |- (fusionWestIn);

\draw[flow] (fusion.east) -- (17.55,-0.35);

\end{tikzpicture}}
\caption{Overview of the proposed quantization-aware vision token pruning framework. Given input visual tokens and the query text, the model computes three complementary signals: group-wise quantization error, global outlier intensity, and semantic relevance. The first two signals are combined into a quantization sensitivity score $\mathbf{S}^{Q}$, which is further fused with the semantic pruning score $\mathbf{S}^{P}$ to produce the final score for selecting the top-$K$ visual tokens.}
\label{fig:method_overview}
\end{figure}

In this section, we detail our unified framework for jointly performing post-training quantization and vision token pruning in MLLMs. Figure~\ref{fig:method_overview} illustrates the overall pipeline. We first formulate the PTQ setting for MLLMs and then describe how quantization sensitivity is incorporated into token selection.

\subsection{Problem Setup and Baseline Framework}
\noindent\textbf{MLLMs Post Training Quantization.} We formulate the quantization of a linear layer $\mathbf{Y}=\mathbf{X}\mathbf{W}$ within the MLLMs as:
$$
\hat{\mathbf{Y}}=Q(\mathbf{X})Q(\mathbf{W}),
$$
where $\mathbf{X} \in \mathbb{R}^{B\times L\times d_{in}}$ denotes the activation tensor, $\mathbf{W} \in \mathbb{R}^{d_{in} \times d_{out}}$ is the weight matrix, and $Q(\cdot)$ represents the quantization function.

Following advanced PTQ frameworks for MLLMs, we employ asymmetric uniform quantization with a group-wise scaling strategy to handle the severe activation outliers inherent in large language models. For a given tensor $\mathbf{T}$ partitioned into groups, its quantized representation $Q(\mathbf{T})$ for a specific group is computed as:
$$
Q(\mathbf{T})=\mathrm{Clamp}(\lfloor\cfrac{\mathbf{T}-\mathbf{Z}}{s}\rceil,0,2^b-1) \cdot s + \mathbf{Z},
$$
where $b$ is the bit-width, $s=\cfrac{max(\mathbf{T})-min(\mathbf{T})}{2^b-1}$ is the scale, and $\mathbf{Z}=min(\mathbf{T})$ is the zero-point.

\vspace{0.3em}\noindent\textbf{Vision Tokens Pruning.} We consider an MLLM consisting of a vision encoder and a language model. Given an image $I$ and a query text $q$, the vision encoder produces $N$ visual tokens $\{\mathbf{v}_i\}_{i=1}^N$, which are projected into the LLM embedding space:
$$
\mathbf{x}_i = g(\mathbf{v}_i) \in \mathbb{R}^{d},
$$
where $g ( \cdot )$ denotes the multimodal projector. Processing all $N$ tokens is expensive due to the increased sequence length and attention/KV-cache cost. Token pruning seeks a subset  $S \subset \{ 1,...,N \}$ with $ |S|= K$ that preserves task performance. 

\subsection{Quantization-Aware Token Sensitivity}
To ensure that the pruning process does not discard tokens critical for the numerical stability of the quantized model, we introduce a hybrid Quantization Sensitivity Score. This score is designed to capture both the fine-grained local detail loss incurred during quantization and the global intensity of activation outliers. We compute this sensitivity for each visual token $\mathbf{v}_i \in \mathbb{R}^D$ (where $D$ is the hidden dimension) by fusing two orthogonal metrics: Group-wise Quantization Simulation and Global Outlier Intensity.

\vspace{0.3em}\noindent\textbf{Group-wise Quantization Simulation.} The first component measures the local reconstruction error introduced by low-bit group-wise quantization. In modern PTQ frameworks, activations are typically partitioned into smaller groups of size $G$ (\textit{e.g.}, $G=128$) to compute localized scaling factors, thereby mitigating the impact of channel-wise outliers.

To simulate this process, we reshape the token features into $M=\cfrac{D}{G}$ groups, such that the $m$-th group of the $i$-th token is denoted as $\mathbf{v}_{i,m} \in \mathbb{R}^G$. Assuming symmetric INT4 quantization (yielding $2^{4-1}=7$ positive quantization levels), the localized scale $s_{i,m}$ and the resulting quantized representation $\hat{\mathbf{v}}_{i,m}$ are computed as:
$$
s_{i,m}=\cfrac{max(|\mathbf{v}_{i,m}|)}{7}, \quad \hat{\mathbf{v}}_{i,m}=\mathrm{Round}(\cfrac{\mathbf{v}_{i,m}}{s_{i,m} + \epsilon}) \cdot s_{i,m}
$$
where $\epsilon$ is a small constant to prevent division by zero. The quantized groups are concatenated back to form $\hat{\mathbf{v}}_i \in \mathbb{R}^D$. The Group-wise Quantization Error for the $i$-th token, denoted as $\mathbf{E}_i$, is then defined as the $L_2$ norm of the reconstruction residual:
$$
\mathbf{E}_i=||\mathbf{v}_i-\hat{\mathbf{v}}_i||_2
$$
Tokens with high $\mathbf{E}_i$ are intrinsically difficult to quantize at a local level and suffer significant information loss, making them critical candidates for retention.

\vspace{0.3em}\noindent\textbf{Global Outlier Intensity.} While the group-wise error captures localized quantization difficulty, it may fail to explicitly penalize the removal of tokens that harbor extreme global outliers. These outlier-bearing tokens dictate the maximum activation bounds across the entire tensor and are crucial for the preservation of emergent properties in large language models.

To explicitly protect these structural outliers, we define the Global Outlier Intensity, $\mathbf{R}_i$, for the $i$-th token as the spread of its activation values across all $D$ channels:
$$
\mathbf{R}_i = \max_{j \in \{1,...,D\}}{(\mathbf{v}_{i,j})} - \min_{j \in \{1,...,D\}}{(\mathbf{v}_{i,j})}
$$
A large $\mathbf{R}_i$ indicates the presence of severe activation outliers, rendering the token highly sensitive to quantization shifts if discarded.

\vspace{0.3em}\noindent\textbf{Hybrid Quantization Sensitivity and Score Fusion.} To construct a comprehensive measure that balances local detail preservation with global outlier protection, we normalize both metrics independently into the $[ 0 , 1 ]$ interval across the $N$ spatial tokens within a batch The final Quantization Sensitivity Score, $\mathbf{S}_i^Q$, is formulated as an equally weighted sum of the normalized metrics:
$$
\mathbf{S}_i^Q = \cfrac{1}{2}\left(\frac{\mathbf{E}_i-min(\mathbf{E})}{max(\mathbf{E})-min(\mathbf{E})}\right) + \cfrac{1}{2}\left(\frac{\mathbf{R}_i-min(\mathbf{R})}{max(\mathbf{R})-min(\mathbf{R})}\right).
$$

Finally, we integrate this quantization sensitivity with the traditional vision tokens pruning method score, $\mathbf{S}_i^P$, to guide the token selection process. We introduce a hyper-parameter $\alpha \in [0,1]$ to control the trade-off between semantic alignment and numerical stability:
$$
\mathbf{S}_i^{Final} = \alpha \mathbf{S}_i^P + (1-\alpha)\mathbf{S}_i^Q.
$$
In this way, our method recalibrates the token selection criteria so that the pruned visual sequence remains both semantically informative for the query and robust to the degradation introduced by low-bit PTQ.

\section{Experiments}

\subsection{Implementation Details}
\noindent\textbf{Models and Datasets.} We conduct our experiments using the LLaVA-1.3~\cite{llava} 7B and 13B models, alongside the LLaVA-1.5~\cite{liu2023improvedllava} 7B model, serving as our representative MLLMs. For evaluation, we primarily utilize the ScienceQA~\cite{scienceqa} benchmark, which necessitates complex multimodal reasoning over integrated diagrams and textual contexts. Accuracy is reported as the primary performance metric.

\vspace{0.3em}\noindent\textbf{Quantization Setup.} We adopt QVLM~\cite{qvlm} as our foundational PTQ framework and CDPruner~\cite{cdpruner} as our baseline for vision token pruning. To simulate realistic low-bit deployment scenarios, we apply PTQ to the LLaVA architecture, utilizing W4A4 quantization (4-bit weights and 4-bit activations) as our default configuration. This represents a particularly challenging regime, as activation outliers can severely degrade model performance at such low precision. We evaluate our proposed approach against these W4A4-quantized baselines, both with and without standard token pruning.

\begin{table}[h]
\centering
\footnotesize 
\caption{Comparisons with quantization methods and vision tokens pruning methods for LLaVA-v1.3 and LLaVA-v1.5 7B models on the ScienceQA dataset. Question classes: NAT = natural science, SOC = social science, LAN = language science, TXT = text context, IMG = image context, NO = no context.}
\label{tab:main_results}
\begin{tabular*}{\textwidth}{@{\extracolsep{\fill}} c|c|c|c|ccc|ccc|l @{}}
\hline
\multirow{2}{*}{} & \multirow{2}{*}{Bits} & \multirow{2}{*}{Method} & Tokens & \multicolumn{3}{c|}{Subject} & \multicolumn{3}{c|}{Context Modality} & \multirow{2}{*}{Average} \\
 & & & & NAT & SOC & LAN & TXT & IMG & NO & \\
\hline
\multirow{6}{*}{\makebox[0.8cm]{\rotatebox{90}{LLaVA-7B}}} 
 & FP & - & 256 & 89.43 & 95.61 & 85.09 & 88.61 & 87.41 & 88.15 & 89.60 \\ \cline{2-11}
 & \multirow{5}{*}{W4A4} & QVLM & 256 & 77.62 & 89.99 & 78.18 & 75.95 & 77.74 & 80.77 & 80.36 \\
 & & With Prune & 128 & 77.62 & 87.63 & 77.73 & 75.90 & 76.45 & 80.77 & 79.75 \\
 & & \textbf{Ours} & 128 & 78.69 & 90.55 & 77.82 & 77.22 & 78.98 & 80.77 & 80.95{\color{red}(+1.20)} \\
 & & With Prune & 32 & 77.58 & 82.34 & 77.73 & 75.76 & 74.07 & 80.77 & 78.61 \\
 & & \textbf{Ours} & 32 & 79.40 & 88.08 & 78.00 & 78.25 & 78.78 & 80.77 & 80.85{\color{red}(+2.24)} \\
\hline
\multirow{6}{*}{\makebox[0.8cm]{\rotatebox{90}{LLaVA-13B}}}
 & FP & - & 256 & 90.99 & 94.94 & 88.18 & 90.03 & 88.45 & 90.73 & 91.09 \\ \cline{2-11}
 & \multirow{5}{*}{W4A4} & QVLM & 256 & 82.90 & 91.34 & 81.82 & 81.82 & 80.91 & 84.60 & 84.39 \\
 & & With Prune & 128 & 82.95 & 89.43 & 82.00 & 82.06 & 80.22 & 84.60 & 84.06 \\
 & & \textbf{Ours} & 128 & 83.08 & 92.13 & 82.09 & 82.16 & 81.61 & 84.60 & 84.72{\color{red}(+0.66)} \\
 & & With Prune & 32 & 82.19 & 84.93 & 81.91 & 81.62 & 77.34 & 84.60 & 82.69 \\
 & & \textbf{Ours} & 32 & 82.99 & 90.33 & 82.00 & 82.26 & 80.66 & 84.60 & 84.27{\color{red}(+1.58)} \\
\hline
\multirow{6}{*}{\makebox[0.8cm]{\rotatebox{90}{LLaVA-1.5}}}
 & FP & - & 576 & 66.30 & 70.75 & 68.45 & 66.23 & 65.49 & 68.50 & 67.79 \\ \cline{2-11}
 & \multirow{5}{*}{W4A4} & QVLM & 576 & 60.12 & 64.00 & 58.91 & 59.38 & 60.04 & 59.16 & 60.62 \\
 & & With Prune & 128 & 60.44 & 62.99 & 59.18 & 59.43 & 60.09 & 59.16 & 60.65 \\
 & & \textbf{Ours} & 128 & 61.06 & 64.12 & 59.27 & 60.56 & 61.33 & 59.16 & 61.24{\color{red}(+0.59)} \\
 & & With Prune & 32 & 60.57 & 64.57 & 59.18 & 60.07 & 60.93 & 59.16 & 61.05 \\
 & & \textbf{Ours} & 32 & 61.06 & 66.14 & 59.00 & 60.61 & 62.07 & 59.16 & 61.59{\color{red}(+0.54)} \\
\hline
\end{tabular*}
\end{table}

\subsection{Main Results on Visual Question Answering}
As shown in Table~\ref{tab:main_results}, our Quantization-Aware Pruning method consistently outperforms the naive combination baseline. For LLaVA 1.3 7B, the results show that directly applying standard vision token pruning to PTQ-optimized MLLMs leads to a clear performance drop. In contrast, our quantization-aware method retains more outlier-rich tokens and better preserves critical visual information. At a 50\% token retention rate, our method improves accuracy by 1.2\% over the naive pruning baseline. This advantage becomes larger as the pruning ratio increases. When retaining only 12.5\% of the original visual tokens, our method improves performance by 2.24\% and even surpasses the dense quantized model, which keeps all visual tokens, by 0.49\%. These results show that our framework effectively co-optimizes vision token pruning and PTQ.

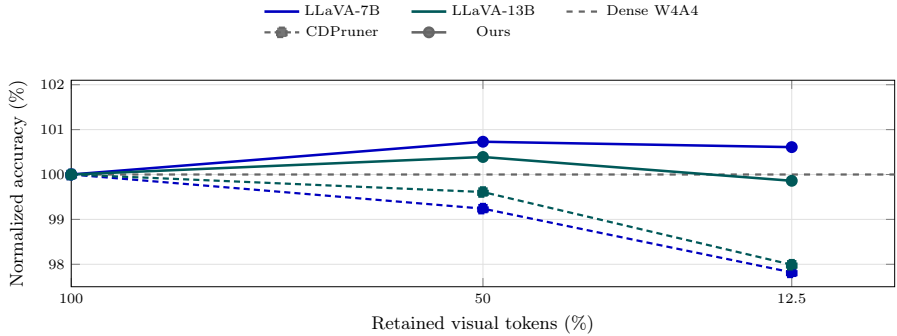
\begin{figure}[t]
\centering
\resizebox{0.98\textwidth}{!}{\begin{tikzpicture}
\begin{axis}[
    width=15.8cm,
    height=5.15cm,
    xmin=0, xmax=100,
    x dir=reverse,
    ymin=97.5, ymax=102.1,
    xtick={100,50,12.5},
    xticklabels={100,50,12.5},
    ytick={98,99,100,101,102},
    xlabel={Retained visual tokens (\%)},
    ylabel={Normalized accuracy (\%)},
    tick label style={font=\scriptsize},
    label style={font=\small},
    legend style={
        draw=none,
        fill=white,
        fill opacity=0.9,
        text opacity=1,
        font=\scriptsize,
        at={(0.5,1.16)},
        anchor=south,
        legend columns=3,
        /tikz/every even column/.append style={column sep=0.45cm}
    },
    grid=both,
    major grid style={draw=gray!24},
    minor grid style={draw=gray!12},
    every axis plot/.append style={line width=1.1pt, mark size=2.3pt},
    clip=false
]

\addplot[black!60, dashed, forget plot] coordinates {(100,100) (0,100)};

\addplot[blue!75!black, densely dashed, mark=square*, forget plot]
coordinates {(100,100) (50,99.24) (12.5,97.82)};
\addplot[blue!75!black, mark=*, line width=1.25pt, forget plot]
coordinates {(100,100) (50,100.73) (12.5,100.61)};

\addplot[teal!70!black, densely dashed, mark=square*, forget plot]
coordinates {(100,100) (50,99.61) (12.5,97.99)};
\addplot[teal!70!black, mark=*, line width=1.25pt, forget plot]
coordinates {(100,100) (50,100.39) (12.5,99.86)};

\addlegendimage{blue!75!black, line width=1.3pt}
\addlegendentry{LLaVA-7B}
\addlegendimage{teal!70!black, line width=1.3pt}
\addlegendentry{LLaVA-13B}

\addlegendimage{black!60, dashed, line width=1.1pt}
\addlegendentry{Dense W4A4}
\addlegendimage{black!60, densely dashed, mark=square*, line width=1.1pt}
\addlegendentry{CDPruner}
\addlegendimage{black!60, solid, mark=*, line width=1.25pt}
\addlegendentry{Ours}

\end{axis}
\end{tikzpicture}}
\vspace{-10pt}
\caption{Normalized accuracy retention versus retained visual-token ratio for LLaVA-7B and LLaVA-13B under W4A4 PTQ. Each curve is normalized by the dense W4A4 baseline of its model. As the token budget shrinks, semantic-only pruning degrades more noticeably, while our method remains consistently closer to, and often above, the dense PTQ baseline.}
\label{fig:pruning_ratio_plot}
\end{figure}

The same conclusion also holds for LLaVA 1.3 13B and LLaVA 1.5 7B. For LLaVA 1.3 13B, our method improves the average accuracy from 84.06\% to 84.72\% at 256$\rightarrow$128 tokens and from 82.69\% to 84.27\% at 256$\rightarrow$32 tokens. For LLaVA 1.5 7B, our method similarly improves the average accuracy from 60.65\% to 61.24\% at 576$\rightarrow$128 tokens and from 61.05\% to 61.59\% at 576$\rightarrow$32 tokens. More importantly, in five out of the six pruned settings across all three models, our method even surpasses the dense W4A4 baseline while using substantially fewer visual tokens, which further demonstrates that it can effectively combine vision token pruning with PTQ. This trend is further illustrated in Fig.~\ref{fig:pruning_ratio_plot} for LLaVA-7B and LLaVA-13B, where our method remains consistently closer to the dense PTQ baseline under lower token budgets.

\subsection{Ablation Study}
In this section, we conduct an ablation study to evaluate the formulation of the proposed quantization sensitivity metric. To systematically quantify this sensitivity, we explore several alternative metrics inspired by established quantization techniques. Drawing on the principles of LLM.int8()~\cite{llmint8} and GPTQ~\cite{GPTQ}, which isolate or measure outlier magnitudes to assess quantization sensitivity, we evaluate the $L_{\infty}$, $L_1$ and $L_2$ norms of the token activations as proxy scores. Furthermore, inspired by the utilization of intra-token variance in Q-VLM~\cite{qvlm}to capture information density and distribution spread, we investigate token variance as an alternative metric. Beyond these heuristic proxies, we also assess methods that explicitly simulate quantization error. Specifically, we compare token-wise absolute maximum (AbsMax) and percentile-clipping quantization simulations. Table~\ref{tab:ablation_method} summarizes the empirical performance of these varying sensitivity metrics, utilizing the LLaVA 1.3 7B model configured to retain 32 visual tokens as the experimental baseline.

\begin{table}[tbp]
\centering
\caption{Ablation study about methods to calculate quantization sensitivity}
\vspace{-8pt}
\label{tab:ablation_method}
\begin{tabular}{c|ccc}
\toprule
\textbf{Methods} & \textbf{G1 Accuracy$\uparrow$} & \textbf{G7 Accuracy$\uparrow$} & \textbf{Average.$\uparrow$} \\
\midrule
- & 80.73 & 74.82 & 78.61 \\
$L_{\infty}$ Norm & 82.42 & 75.81 & 80.05 \\
Token-wise AbsMax & 82.42 & 76.07 & 80.15 \\
Variance & 82.67 & 75.68 & 80.17 \\
$L_1$ Norm & 82.67 & 76.27 & 80.38 \\
$L_2$ Norm & 82.93 & 76.20 & 80.52 \\
Clip AbsMax & 83.00 & 76.14 & 80.55 \\
Group-wise AbsMax & 82.97 & 76.86 & 80.78 \\
Outlier Intensity & \textbf{83.11} & 76.60 & 80.78 \\
\rowcolor{blue!5} 
Combine & 82.97 & \textbf{77.06} & \textbf{80.85} \\
\bottomrule
\end{tabular}
\end{table}

As demonstrated in Table~\ref{tab:ablation_method}, the integration of any quantization sensitivity metric consistently yields superior performance compared to the semantic-only baseline. Notably, the group-wise AbsMax simulation (80.78\%) closely models actual deployment conditions, thereby accurately capturing local detail loss. While the global outlier intensity metric alone achieves a highly competitive 80.78\% by protecting extreme activation bounds, our proposed combined approach, which integrates simulated group-wise error with outlier intensity, attains the highest average accuracy of 80.85\%. These results confirm that balancing fine-grained detail preservation with structural outlier protection constitutes the most robust pruning strategy for low-bit inference.

\section{Conclusion}

In this work, we identified and addressed the non-trivial performance degradation that occurs when integrating vision token pruning with PTQ in MLLMs. 
Our investigations revealed that standard semantic pruning often discards outlier-rich tokens. Although these tokens may appear less important semantically, they are structurally important for maintaining numerical stability in low-bit quantization.
To overcome this limitation, we introduced a unified Quantization-Aware Pruning framework driven by a novel hybrid quantization sensitivity metric. By simultaneously capturing local detail loss through simulated group-wise error and global extreme values via outlier intensity, our method provides a comprehensive assessment of token-level quantization difficulty. 
Experimental results across representative LLaVA architectures and VQA benchmarks validate the superiority of our approach. 
By explicitly retaining quantization-critical tokens, our framework not only mitigates the severe accuracy degradation typical of low-bit regimes (\textit{e.g.}, W4A4) but also leverages high pruning ratios to surpass the performance of dense quantization baselines. This work establishes a highly effective methodology for co-optimizing token pruning and quantization, paving the way for the robust and efficient deployment of advanced MLLMs.


%
%
%
%
\bibliographystyle{splncs04}
\bibliography{main}





\end{document}